 \documentclass[pmlr,twocolumn,10pt]{jmlr} 

\usepackage{booktabs}
\usepackage{siunitx}

\usepackage{multirow}
\usepackage{multicol}
\usepackage{pifont}

\usepackage{listings}
\usepackage{lipsum} 

\usepackage[switch]{lineno}

\theorembodyfont{\upshape}
\theoremheaderfont{\scshape}
\theorempostheader{:}
\theoremsep{\newline}

\jmlrworkshop{Machine Learning for Health (ML4H) 2024} 

 \title[Synthetic Data Generation for Clinical QA]{Give me Some Hard Questions: Synthetic Data Generation \\ for Clinical QA}

\author{%
\Name{Fan Bai} \Email{fbai3@jh.edu}\\
\Name{Keith Harrigian} \Email{kharrigian@jhu.edu}\\
\addr Johns Hopkins University, Baltimore, MD, USA
\AND
\Name{Joel Stremmel}\thanks{Work done while at Optum.} \Email{joel.stremmel@thomsonreuters.com}\\
\addr Thomson Reuters, Eagan, MN, USA 
\AND
\Name{Hamid Hassanzadeh} \Email{hamid.hassanzadeh@optum.com}\\
\Name{Ardavan Saeedi} \Email{ardavan.saeedi@optum.com}\\
\addr Optum, Minnetonka, MN, USA 
\AND
\Name{Mark Dredze} \Email{mdredze@cs.jhu.edu}\\
\addr Johns Hopkins University, Baltimore, MD, USA
}

\newcommand{\PreserveBackslash}[1]{\let\temp=\\#1\let\\=\temp}
\newcolumntype{C}[1]{>{\PreserveBackslash\centering\arraybackslash}p{#1}}
\newcolumntype{R}[1]{>{\PreserveBackslash\raggedleft\arraybackslash}p{#1}}
\newcolumntype{L}[1]{>{\PreserveBackslash\raggedright\arraybackslash}p{#1}}

\usepackage{colortbl} 

\begin{document}

\maketitle

\begin{abstract}
Clinical Question Answering (QA) systems enable doctors to quickly access patient information from electronic health records (EHRs). However, training these systems requires significant annotated data, which is limited due to the expertise needed and the privacy concerns associated with clinical data. This paper explores generating Clinical QA data using large language models (LLMs) in a zero-shot setting. We find that naive prompting often results in easy questions that do not reflect the complexity of clinical scenarios. To address this, we propose two prompting strategies: 1) instructing the model to generate questions that do not overlap with the input context, and 2) summarizing the input record using a predefined schema to scaffold question generation. Experiments on two Clinical QA datasets demonstrate that our method generates more challenging questions, significantly improving fine-tuning performance over baselines. We compare synthetic and gold data and find a gap between their training efficacy resulting from the quality of synthetically generated answers.
\end{abstract}
\begin{keywords}
Clinical QA, Synthetic Data Generation, Large Language Model
\end{keywords}


\paragraph*{Data and Code Availability} We use two Clinical QA datasets: RadQA\footnote{\url{https://physionet.org/content/radqa/1.0.0/}} \citep{soni-etal-2022-radqa} and MIMIC-QA\footnote{\url{https://physionet.org/content/mimic-iii-question-answer/1.0.0/}} \citep{mimicqa}, both of which can be accessed through PhysioNet. Our code is available on Github.\footnote{\url{https://github.com/bflashcp3f/synthetic-clinical-qa}}



\paragraph*{Institutional Review Board (IRB)} Our research does not involve human subjects and therefore does not require IRB approval.

\section{Introduction}
\label{sec:intro}

\begin{figure}[!t]
\floatconts
    {fig:motivation}
    {\caption{As shown by the text highlighted in matching colors, directly prompting LLMs to generate questions from clinical records often results in simple questions that overlap with the input context and can be easily answered through superficial matching.}}
    {\includegraphics[width=0.45\textwidth]
  {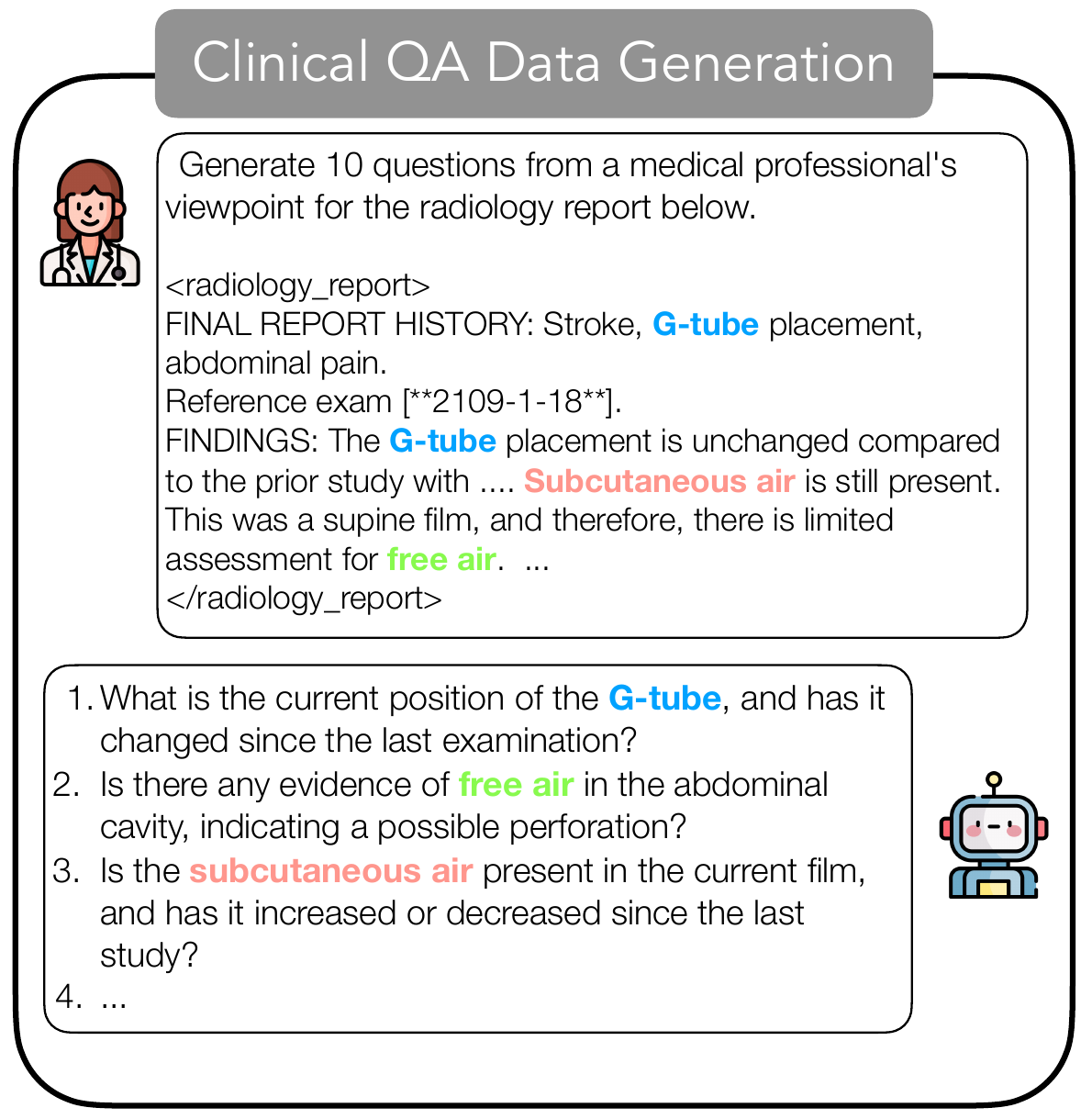}}
\end{figure}

As electronic health records (EHRs) expand to include more comprehensive patient information, obtaining timely access to critical details becomes increasingly challenging. Clinical Question Answering (QA) systems \citep{PATRICK2012292, pampari-etal-2018-emrqa} can transform how doctors interact with patient information by providing precise answers to specific questions. While advances in document-level QA systems can benefit Clinical QA, the nuances of clinical terminology and medical context necessitate training specialized systems. If a doctor asks, ``Did the cardiac silhouette enlarge?'' the system must understand that this refers to the size of the heart and either provide an answer from the context or indicate that the question is unanswerable \citep{soni-etal-2022-radqa}.

Unlike developing QA systems for general domains, where constructing large-scale annotated resources is feasible \citep{rajpurkar-etal-2016-squad}, the development of Clinical QA systems is constrained by the scarcity of annotated data. Collecting such data is challenging due to the cost and expertise required for accurate annotation, as well as the privacy concerns associated with creating shared medical record datasets \citep{johnson2016mimic}. Consider the current state of Clinical QA datasets: even with restrictive licenses, few datasets are available, and many of these contain data artifacts due to how they were constructed (\S\ref{sec:data}).

These challenges incentivize the development of automated methods for creating synthetic Clinical QA datasets. Ideally, automatic data generation should: 1) create high-quality, realistic questions and 2) provide relevant questions that cannot be answered from the input documents (unanswerable questions). Previous methods have struggled to achieve both of these goals. Question generation has relied on templates \citep{pampari-etal-2018-emrqa}, which tend to produce formulaic questions, or has depended on extensive data annotation \citep{du-etal-2017-learning, alberti-etal-2019-synthetic, puri-etal-2020-training}. Similarly, unanswerable questions are often generated by artificially tweaking gold answerable questions \citep{gautam-etal-2023-lightweight}, limiting the scope of these examples.

In this paper, we create high-quality Clinical QA data using instruction-tuned large language models (LLMs) to address both of these challenges. Leveraging the strong zero-shot performance of recent LLMs, we adopt a distillation approach: first, we instruct LLMs to generate questions based on the input document, then distill answers from the LLMs, with unanswerable questions naturally emerging during the second step. However, we observe that naive prompting often produces questions that are too basic to train effective Clinical QA systems, as these questions tend to incorporate specific phrasing from the input document and thus can be answered through superficial lexical matching.

To facilitate LLMs generating better questions, we explore two prompting strategies: 1) designing targeted instructions, such as instructing the model to generate questions that do not overlap with the input context, and 2) introducing a summarization step to exclude extraneous details and scaffold question generation. We demonstrate the effectiveness of our proposed methods on two Clinical QA datasets: RadQA \citep{soni-etal-2022-radqa} and MIMIC-QA \citep{mimicqa}, using two LLMs: \texttt{Llama3-8B} \citep{llama3modelcard} and \texttt{GPT-4o} \citep{achiam2023gpt}. Empirical results show that our generated data leads to significant improvements compared to existing question generation methods when used to fine-tune task-specific models. Further analyses show that our methods generate more challenging questions, such as those that require a deeper understanding of the input context and cannot be answered through superficial matching.

Moreover, we examine the differences between synthetic and gold data to identify the remaining limitations of synthetic data. We conduct a controlled comparison by introducing a new setting where LLMs generate answers for gold questions and then use these pairs for fine-tuning. We find that when both use synthetic answers, the performance gap between synthetic and gold questions narrows as the number of documents increases. However, the gap persists with gold answers, highlighting that synthetic answer quality remains a challenge for this task.

\section{Clinical QA Corpora Generation using LLMs}

For Clinical QA corpora generation, the goal is to create multiple QA pairs for each input document, covering both answerable and unanswerable questions, as doctors may seek information that is unavailable in the context. Here we do not assume access to labeled data, a practical setting for clinical domains, and thus adopt a distillation approach: we first generate questions using LLMs and then distill answers from them, with unanswerable questions emerging during the second step. The challenge lies in generating questions that closely mimic real clinical scenarios.

For the radiology report in Figure \ref{fig:motivation}, an LLM could generate the following QA pair:
\begin{quote}
{\bf Q}: What is the assessment of air?\\
{\bf A}: There is limited assessment for free air.    
\end{quote}

While correct, this QA pair is too easy because the question and answer have high word overlap, and a doctor is unlikely to use the exact terminology that appears in an unseen record. Instead, the question ``Is there any evidence of gastrointestinal perforation?'' would make for a more informative and challenging QA pair, as it requires the system to understand that ``gastrointestinal perforation'' means a hole has formed in the intestine and may present as free air in the abdomen. We seek a process for generating these types of questions.

\subsection{Methods}
We propose two prompting strategies to guide LLMs in generating better clinical questions. We then use these generated QA pairs to train a supervised Clinical QA system, where better synthetic training data leads to better performance.

\paragraph{Targeted Instruction} Recent LLMs are trained to follow human language instructions \citep{ouyang2022training}. In our case, since we aim to generate questions that cannot be answered through superficial matching, we demonstrate that this can be achieved by instructing the model to generate questions that \textit{``do not contain any words from the input context"} (referred to as ``No Overlap"). As confirmed in our analysis in \S\ref{sec:results}, while this instruction does not eliminate overlapping questions, it significantly increases the proportion of questions that are more challenging and informative, thereby improving Clinical QA model training.

\paragraph{Input Summarization} To guide LLMs toward more relevant content and away from extraneous details, we formulate question generation as a two-step process: summarization and question generation. The summarization step abstracts the input context into a structured summary using a predefined schema, which includes high-level attributes covering the essential aspects a doctor would consider when formulating questions. For instance, in radiology reports, attributes such as \textit{``symptoms"} and \textit{``areas examined"} are included. We use a template-based approach to create an attributed summary, prompting LLMs to generate a summary according to a predefined JSON-formatted schema \citep{bai2024schemadriven}. Our results demonstrate that this method complements the proposed ``No Overlap" instruction, as it further promotes the generation of challenging questions and improves the diversity of questions generated within each document. The prompts used for summarization, along with other methods like answer distillation, can be found in Appendix \S\ref{sec:full_prompt}. Appendix \S\ref{sec:att_sum} provides an example of the generated summary and validate the usefulness of pre-defined schema in summarization through ablation studies.

\section{Experimental Setup}

\begin{table}[t!]

\floatconts
    {tab:data_stat}
    {\caption{Data statistics of RadQA and MIMIC-QA.}}
    {\begin{center}
    \setlength{\tabcolsep}{5pt}
    
        \scalebox{0.8}{
        \begin{tabular}{lcc}
        
        \toprule
        
         & \textbf{RadQA} & \textbf{MIMIC-QA}\\
        & \citep{soni-etal-2022-radqa} & \citep{mimicqa}\\
        
        \midrule 
        doc. source & radiology reports & clinical notes \\
        \# docs & 1009 & 36 \\
        doc split (tr/dv/ts) & 803 / 102 / 104 & 0 / 0 / 36 \\
        \# questions & 3074 & 1287 \\ 
        \# ques. / doc & 3.0 & 35.8 \\ 
        \# ave. doc length & 71.2 & 1875.6 \\
        \# ave. ques. length & 7.2 & 8.7 \\
        \# ave. ans. length & 15.6 & 11.1 \\
        unanswerable ques. & \ding{51} & \ding{55}\\
        
        \bottomrule 
        \end{tabular}
        }
    \end{center}}
\end{table}

\subsection{Data}
\label{sec:data}

We experiment with two Clinical QA datasets: RadQA \citep{soni-etal-2022-radqa} and MIMIC-QA \citep{mimicqa}. Both datasets source their documents from MIMIC-III \citep{johnson2016mimic}. Each dataset unit consists of a clinical document associated with a set of QA pairs, where answers are text spans extracted from the context (extractive QA). Dataset statistics are presented in Table \ref{tab:data_stat}.

\paragraph{RadQA} contains questions about radiology reports, comprising 1009 reports from 100 sampled patients, with questions and answers manually annotated by physicians. The questions are derived from the clinical referral section of the reports, which outlines the reason for the radiology test from the ordering physicians. This approach minimizes biases that could arise from annotators seeing the answer context and allows for the creation of unanswerable questions, as annotators are unaware of the specific content in the report. As discussed below, we consider this to be the highest-quality Clinical QA dataset available.

\paragraph{MIMIC-QA} contains only a test set, which includes 36 MIMIC-III discharge summaries and 1287 associated QA pairs. Among these questions, 975 are automatically generated and later verified by medical students, while they directly author 312 after reading the full documents. This process results in easier, overlapping questions, with no unanswerable ones. Following \citet{mimicqa}, we use the training set of emrQA \citep{pampari-etal-2018-emrqa}, which contains similar MIMIC-III notes, to generate QA pairs.

The challenges of obtaining real clinical data mean that only two Clinical QA datasets are suitable for our study, underscoring the need for synthetic data. Table~\ref{tab:qa_dataset} in the appendix describes ten existing Clinical QA datasets, of which only four are publicly available (a common problem for clinical data). Two of these four aren't useful for our study: emrQA \citep{pampari-etal-2018-emrqa} was created automatically via template filling, resulting in questions that lack diversity and realism. DiSCQ \citep{lehman-etal-2022-learning} contains human-authored questions but does not provide gold answers. We evaluate using the two remaining datasets: RadQA and MIMIC-QA. We include MIMIC-QA despite serious limitations: it lacks a training set, does not include unanswerable questions, and most of its questions are machine-generated based on answers. Nevertheless, we include it as our best option for a second dataset.

\subsection{Baselines}

We compare our methods with existing question generation approaches that have been validated on general domain datasets.\footnote{For MIMIC-QA \citep{mimicqa}, we do not directly compare our results with results in the original paper because they use the same question generation model for both training and (the majority of) evaluation data, leading to identical question distributions in both sets. This would create an unfair advantage over our results, where LLMs operate in a zero-shot setting.} The specific prompts for each baseline are presented in Appendix \S\ref{sec:full_prompt}.

\paragraph{Direct Instruction} We instruct the model to generate questions directly from the input context using task descriptions. For example, \textit{``Considering the radiology report provided above, generate 10 questions from a medical professional's viewpoint."}

\paragraph{Temperature Annealing \citep{viswanathan-etal-2023-prompt2model}} When running Direct Instruction, we anneal the temperature of nucleus sampling from 0 to 1 during question generation to promote diversity. For other methods, we set the temperature to 0 (i.e., greedy decoding) to increase result reproducibility.

\paragraph{Question Prefix \citep{yadav-etal-2024-explicit-implict}} We incorporate explicit instructions to generate questions with varied prefixes. For example, \textit{`Please ensure each question starts with a different prefix, such as ``is," ``does," ``has," ``which," ``what," ``how," and ``where".'} Our analysis confirms that this instruction enhances the diversity of question types.

\subsection{LLMs}
We experiment with two LLMs: \texttt{Llama3-8B-Instruct} \citep{llama3modelcard} and \texttt{GPT-4o}\footnote{\texttt{gpt-4o-2024-05-13} version.} \citep{achiam2023gpt}. In compliance with the licensing terms of the datasets, we use a HIPAA-compliant version of \texttt{GPT-4o}, provided through the Microsoft Azure API. For \texttt{Llama3-8B}, we deploy it using our in-house A100 GPUs. In preliminary experiments, we also evaluated other medical-specialized LLMs, such as \texttt{Meditron-7B} \citep{chen2023meditron70b}, but these models performed significantly worse, likely due to the lack of proper instruction tuning. As a result, we did not pursue these models further.

\subsection{Synthetic Data Creation}

We randomly sample 64 documents from each dataset for synthetic data generation. As demonstrated in our scaling analysis in \S\ref{sec:results}, this sample size is sufficient for training a competitive QA system and comparing different methods. For RadQA, following the original data creation process, we generate 5 questions per document and then prompt the LLMs to extract answers from the ``Findings" and ``Impression" subsections separately, resulting in 640 synthetic QA pairs per approach. As mentioned in \S~\ref{sec:data}, for MIMIC-QA we use the training set documents from emrQA to generate synthetic QA pairs. Due to the length of the documents (often exceeding 3000 tokens), we employ the ``POS prompting'' method from \citet{yadav-etal-2024-explicit-implict} to split the documents into 500-word segments, generating five questions per segment. This leads to 169 segments and 845 QA pairs. Since MIMIC-QA only contains answerable questions, we apply the same pipeline to over-generate QA pairs and filter out unanswerable ones to ensure five questions per segment.\footnote{Regarding the computational costs of synthetic data generation, \texttt{GPT-4o} on Azure charges \$0.005 per 1,000 input tokens and \$0.015 per 1,000 output tokens as of August 15th, 2024. For 64 RadQA documents, the three steps in our proposed method, i.e., summarization, question generation, and answer distillation, cost approximately \$0.23, \$0.22, and \$0.81, respectively.}

\begin{table}[t!]
\centering
\floatconts
    {tab:radqa_results}
    {\caption{RadQA test set results, where models are fine-tuned on 640 synthetic QA pairs generated from 64 sampled documents. Along with F\textsubscript{1}, we include Reference Overlap (RO), which considers a predicted answer correct if its character-level index in the input document overlaps with the gold answer's index. Results are averaged across three random seeds, with standard deviations indicated as subscripts. Our proposed ``Summarization + No Overlap" performs best in both F\textsubscript{1} and RO.}}
    {\scalebox{0.8}{
    \begin{tabular}{lcccccccccccc}
    \toprule
     
    & \multicolumn{2}{c}{\texttt{Llama3-8B}} & \multicolumn{2}{c}{\texttt{GPT-4o}} \\
    
    \cmidrule(l){2-3}  \cmidrule(l){4-5}  
    
    \textbf{Method} & F\textsubscript{1} & RO & F\textsubscript{1} & RO \\
    
    \midrule
    
    Direct Instruct. & 52.0\textsubscript{1.1} & 61.7\textsubscript{3.7} & 60.6\textsubscript{1.8} & 70.8\textsubscript{2.7} \\
    
    Temp. Anneal. \citeyearpar{viswanathan-etal-2023-prompt2model} & 51.3\textsubscript{3.5} & 60.9\textsubscript{5.5} & 55.9\textsubscript{3.7} & 65.1\textsubscript{4.7}  \\
    
    Ques. Prefix \citeyearpar{yadav-etal-2024-explicit-implict} & 54.9\textsubscript{1.6} & 65.8\textsubscript{1.6} & 57.4\textsubscript{1.8} & 66.2\textsubscript{1.3}  \\
    
    No Overlap & 57.8\textsubscript{1.1} & 70.6\textsubscript{1.5} & 63.7\textsubscript{1.9} & 75.4\textsubscript{2.6}   \\

    Sum. + Direct Instru. & 55.1\textsubscript{1.3} & 65.5\textsubscript{2.7} & 59.6\textsubscript{1.3} & 71.1\textsubscript{2.1}   \\

    
     Sum. + No Overlap & \textbf{60.8}\textsubscript{2.3} & \textbf{74.2}\textsubscript{2.3} & \textbf{65.0}\textsubscript{1.2} & \textbf{76.8}\textsubscript{2.9} \\

    \bottomrule
    \end{tabular}
    }}
\end{table}

\subsection{Fine-tuning Clinical QA Model}
We fine-tune a supervised Clinical QA system using the generated QA pairs. We select BioClinRoBERTa\footnote{\url{https://github.com/facebookresearch/bio-lm}} \citep{lewis-etal-2020-pretrained} as the fine-tuning backbone, a RoBERTa \citep{liu2019roberta} model further pre-trained on medical and clinical corpora, such as PubMed and MIMIC-III. This model has shown strong performance on various clinical tasks, including RadQA. We employ the standard extractive QA fine-tuning architecture, where the document and question are concatenated as input, and the model predicts the start and end positions of the answer span. The model is fine-tuned for 40 epochs with a learning rate of $1\times10^{-5}$ and a batch size of 8 on A100 GPUs. To mitigate the impact of random initialization, we fine-tune each model with three different random seeds and report the average performance.

\subsection{Metrics}
There are two standard metrics for extractive QA: Exact Match (EM) and F1 score. F1 score rewards systems for identifying more words within the answer, while Exact Match only credits systems for getting the answer span exactly right. Since we operate in a zero-shot setting, where no answer span annotation is available to learn about answer boundaries, Exact Match is a less informative metric. Thus, we focus on F1 scores (see EM results in Table \ref{tab:main_results} in the Appendix). Additionally, Clinical QA systems can highlight portions of a record with relevant evidence, rather than only extracting the answer. Therefore, we introduce a less boundary-sensitive metric called Reference Overlap (RO), where an answer is considered correct if its character index in the document overlaps the gold answer's index. While our results show that RO yields similar model rankings to F1, it provides a more realistic evaluation of system performance in a clinical setting.

\begin{table}[t!]
\centering
\floatconts
    {tab:mimicqa_results}
    {\caption{MIMIC-QA test set results, where models are fine-tuned on 845 synthetic QA pairs generated from 64 sampled documents. ``Summarization + Question Prefix" performs best, as most questions are overlapping questions, with question prefix diversity encouraged during dataset creation.}}
    {\scalebox{0.8}{
    \begin{tabular}{lcccccccccccc}
    \toprule

    & \multicolumn{2}{c}{\texttt{Llama3-8B}} & \multicolumn{2}{c}{\texttt{GPT-4o}} \\
    
    \cmidrule(l){2-3}  \cmidrule(l){4-5}
    
    \textbf{Method} & F\textsubscript{1} & RO & F\textsubscript{1} & RO \\
    
    \midrule
    
    Direct Instruct. & 56.4\textsubscript{4.2} & 58.0\textsubscript{3.5} & 51.0\textsubscript{1.3} & 54.8\textsubscript{2.9} \\
    
    Temp. Anneal. \citeyearpar{viswanathan-etal-2023-prompt2model} & 58.4\textsubscript{2.5} & 59.5\textsubscript{1.4} & 47.7\textsubscript{5.3} & 51.5\textsubscript{6.4} \\
    
    Ques. Prefix \citeyearpar{yadav-etal-2024-explicit-implict} & 57.0\textsubscript{4.3} & 57.0\textsubscript{5.4} & 55.4\textsubscript{0.4} & 57.2\textsubscript{1.3} \\
    
    No Overlap & 48.8\textsubscript{3.8} & 51.8\textsubscript{2.2} & 51.6\textsubscript{1.5} & 54.2\textsubscript{3.1}  \\

    Sum. + Direct Instru. & 56.8\textsubscript{0.6} & 57.1\textsubscript{0.7} & 54.5\textsubscript{2.9} & 54.6\textsubscript{4.2}   \\
    
    
    Sum. + Ques. Prefix & \textbf{58.7}\textsubscript{2.0} & \textbf{61.3}\textsubscript{1.9} & \textbf{57.1}\textsubscript{1.9} & \textbf{58.5}\textsubscript{2.2} \\

    \bottomrule
    \end{tabular}
    }}
\end{table}

\begin{table*}[th!]
\floatconts
    {tab:diversity}
    {\caption{Analysis on gold and synthetic questions with four metrics (three questions per document): question length, vocab. size, average pairwise similarity (APS), and average number of unique question prefixes (AQP) per document. We also report the proportion of four question types based on overlap conditions (``\textbf{O}verlap" or ``\textbf{N}o \textbf{O}verlap") and answerability (``\textbf{A}ns." or ``\textbf{U}nans."). Our proposed ``Summarization + No Overlap" method generates more challenging non-overlapping, but answerable (``N.O./A.") questions, approaching the level of gold data.}}
    {\scalebox{0.8}{
    \begin{tabular}{lcccccccc}
    
    \toprule
    & & & & & \multicolumn{4}{c}{\textbf{Question Type (\%)}} \\

    \cmidrule(l){6-9}
    
    & \textbf{Len.} & \textbf{Vocab} $\uparrow$ & \textbf{APS} $\downarrow$ & \textbf{AQP} $\uparrow$ & \textbf{O./A.} & \textbf{O./U.} & \textbf{N.O./A.} & \textbf{N.O./U.} \\
    \midrule
    
    Gold Questions & 7.2 & 312 & 0.445 & 2.6 & 43.8 & 3.7 & 29.5 & 23.0 \\
    \midrule

    Direct Instruct. & 14.3 & 548 & 0.648 & 2.6 & 57.8 & 15.6 & 4.2 & 22.4 \\
    Temp. Anneal. \citeyearpar{viswanathan-etal-2023-prompt2model} & 14.1 & 559 & 0.574 & 2.6 & 56.8 & 19.8 & 3.1 & 20.3 \\
    Quest. Prefix \citeyearpar{yadav-etal-2024-explicit-implict} & 13.8 & 526 & 0.573 & \textbf{3.0} & 50.8 & 22.4 & 3.9 & 22.9 \\
    No Overlap & 11.5 & 353 & 0.555 & 2.6 & 43.2 & 10.9 & 20.3 & 25.5 \\
    Sum. + Direct Instruct. & 17.0 & \textbf{610} & \textbf{0.399} & 2.6 & 52.0 & 24.1 & 5.2 & 18.6\\
    Sum. + No Overlap & 12.0 & 242 & 0.490 & 2.7 & 35.2 & 10.2 & 24.2 & 30.5 \\
    
    \bottomrule 
    \end{tabular}
    }}

\end{table*}


\section{Results and Analyses}
\label{sec:results}


Table \ref{tab:radqa_results} presents test set results on RadQA, where models are fine-tuned on 640 QA pairs generated from 64 sampled documents. We observe that Direct Instruction achieves competitive performance, particularly with \texttt{GPT-4o}, highlighting the strong instruction-following capabilities of recent LLMs. While diversity-promoting approaches can enhance performance, their effectiveness varies depending on the LLM used, e.g., the ``Question Prefix" baseline improves upon Direct Instruction with \texttt{Llama3-8B} but not with \texttt{GPT-4o}. Our proposed ``No Overlap" consistently outperforms other baselines. Combining it with ``Summarization" further boosts performance, achieving the best results in both F\textsubscript{1} and RO. With \texttt{Llama3-8B}, the combined method surpasses Direct Instruction by 8.8 points in F\textsubscript{1} and 12.5 points in RO. With \texttt{GPT-4o}, it achieves improvements of 4.4 points in F\textsubscript{1} and 6.0 points in RO.

Table \ref{tab:mimicqa_results} shows MIMIC-QA results. ``No Overlap" does not work as well as on RadQA, performing worse than ``Question Prefix" with both LLMs. This is due to dataset creation artifacts in MIMIC-QA, where questions are mostly model-generated, whereas the question generation model is trained to promote question-prefix diversity. Combining ``Summarization" with ``Question Prefix" yields the best performance.\footnote{Although 98.8\% of the questions in MIMIC-QA are overlapping questions, we find that recent LLMs do not perform well on this dataset in the zero-shot setting. For example, \texttt{GPT-4o} achieves 58.5 F\textsubscript{1} and 66.7 RO scores. This is because LLMs are not aware of the answer annotation artifacts. When two similar spans, such as patient symptoms, are present in both metadata and the main text, the one in the metadata tends to be annotated as the gold answer, but LLMs may predict the one in the main text.}

\begin{table}[t!]
\centering
\floatconts
    {tab:decomposition}
    {\caption{RadQA test set RO results on four question type subsets based on 1) the overlap condition with the input context and 2) answerability. Our proposed method outperforms baselines significantly on the challenging ``N.O./A." questions.}}
    {\scalebox{0.72}{
    \begin{tabular}{lcccccccccccc}
    \toprule
    
     & \textbf{O./A.} & \textbf{O./U.} & \textbf{N.O./A.} & \textbf{N.O./U.} \\
     \textbf{Method} & (46.4\%) & (7.0\%) & (28.5\%) & (18.1\%) \\
    
    \midrule
    
    Direct Instruct. & 72.9\textsubscript{4.7} & 58.1\textsubscript{9.3} & 55.6\textsubscript{4.3} & 94.3\textsubscript{0.5} \\
    
    Temp. Anneal. \citeyearpar{viswanathan-etal-2023-prompt2model} & 68.2\textsubscript{3.9} & \textbf{62.8\textsubscript{4.0}} & 40.4\textsubscript{10.7} & \textbf{97.3\textsubscript{0.9}} \\
    
    Ques. Prefix \citeyearpar{yadav-etal-2024-explicit-implict} & 67.8\textsubscript{1.3} & 58.9\textsubscript{7.5} & 49.0\textsubscript{3.7} & 91.9\textsubscript{1.8} \\
    
    No Overlap & 78.7\textsubscript{2.7} & 52.7\textsubscript{8.2} & 67.8\textsubscript{4.4} & 87.7\textsubscript{0.5} \\

    Sum. + Direct Instruct. & 75.3\textsubscript{2.7} & 59.7\textsubscript{1.3} & 52.6\textsubscript{3.0} & 94.0\textsubscript{0.5} \\
    
     Sum. + No Overlap & \textbf{79.2\textsubscript{5.8}} & 52.7\textsubscript{4.8} & \textbf{72.6\textsubscript{5.1}} & 86.8\textsubscript{6.3} \\

    \bottomrule
    \end{tabular}
    }}
\end{table}

\paragraph{Performance Decomposition} To gain deeper insights into the performance improvements, we split the RadQA test set into four subsets based on two factors: 1) whether the question overlaps with the input context (after removing stop words) and 2) whether the question is answerable. The percentages of each subset and the RO scores of different methods are presented in Table \ref{tab:decomposition}. We observe that non-overlapping but answerable (``N.O./A.") questions make up 28.5\% of the test set. Our proposed ``Summarization + No Overlap" method outperforms other methods significantly on this subset, contributing to the best overall performance.

\paragraph{Question Analysis}
We analyze the gold and synthetic questions (generated by \texttt{GPT-4o}) on 64 RadQA documents, generating three questions per document to match the gold data. Besides average question length, we assess the questions using three metrics \citep{yu2023large}: vocabulary size, average number of unique question prefixes (AQP) per document, and average pairwise similarity (APS) based on cosine similarity of embeddings (from OpenAI's \texttt{text-embedding-3-small}) between two questions from the same document. We also report the proportion of four question types based on answerability and overlap conditions.

\begin{figure}[tbp]
\floatconts
  {fig:scale-up_ro}
  {\caption{RadQA test set RO results from scaling experiments (8 to 803 documents, 5 to 20 QA pairs per document generated by ``Summarization + No Overlap" with \texttt{GPT-4o}). Fine-tuned models achieve competitive performance with 64 documents and 5 QA pairs per document. Increasing the number of QA pairs per document is beneficial only in low-resource settings (i.e., 8 or 16 documents).}}
  {\includegraphics[width=1\linewidth]{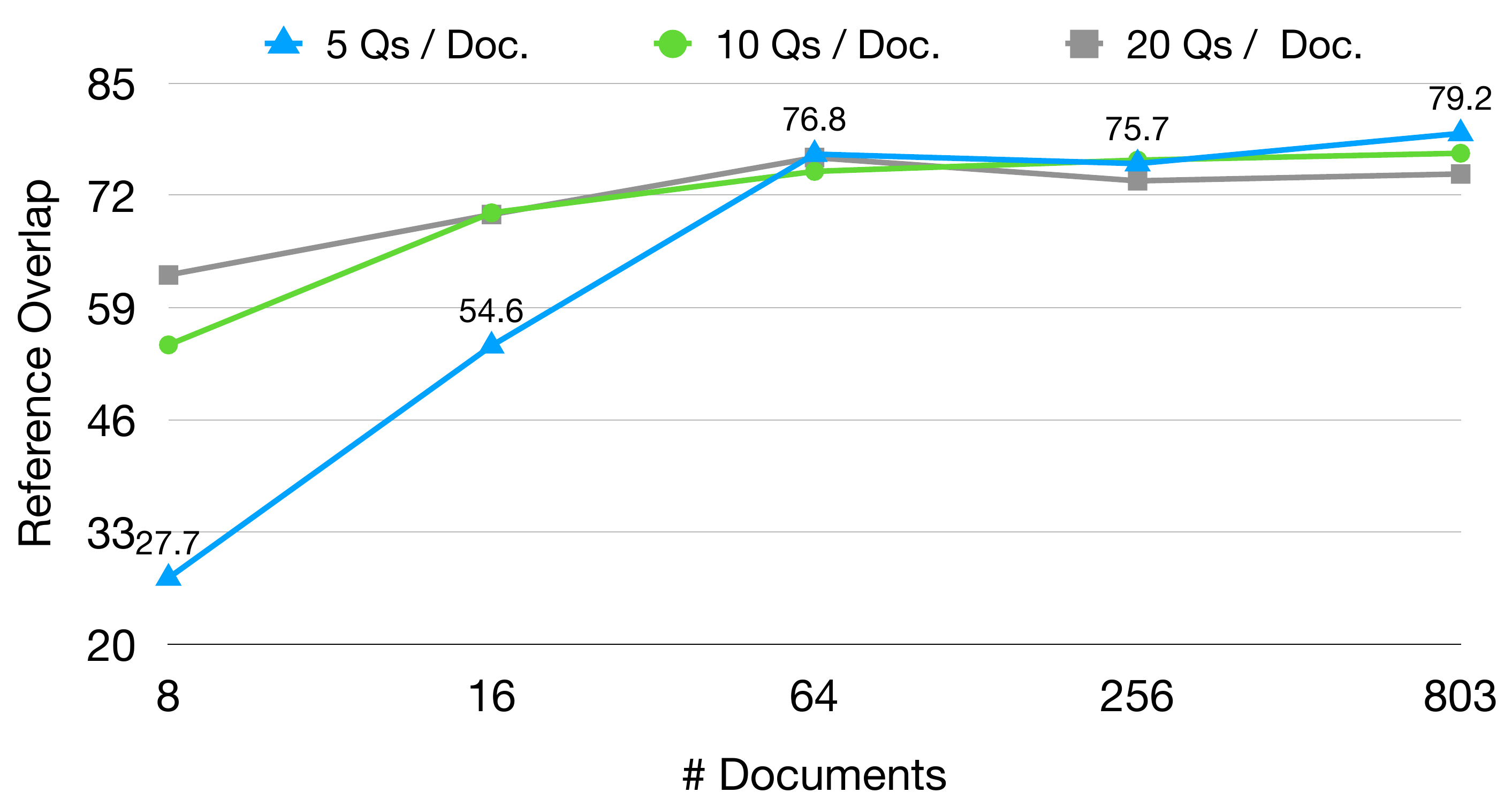}}
\end{figure}

Table \ref{tab:diversity} presents the analysis results. Gold questions are shorter than synthetic ones, and this stylistic difference may contribute to the performance gap. Baselines, such as ``Temperature Annealing," improve linguistic diversity over Direct Instruction by increasing vocabulary, but do not alter the distribution of the four question types. In contrast, ``No Overlap" significantly boosts the percentage of non-overlapping but answerable questions, a crucial factor for training Clinical QA systems. ``Summarization + Direct Instruction" maintains the question distribution but achieves the lowest APS, indicating diversity in the generated data. Combining both methods results in the highest rate of non-overlapping but answerable questions with low APS, approaching the level of gold data.

Table \ref{tab:examples} (Appendix) shows the gold and generated QA pairs for a radiology report. It confirms the succinct style of human-authored questions. Baselines tend to generate questions that closely relate to the context details with direct lexical overlap (highlighted in bold), while our approach generates more challenging questions (introducing relevant but unseen concepts, highlighted in red) that require a deeper understanding of the context to answer.

\begin{figure}[tbp]
\floatconts
  {fig:gold_question}
  {\caption{RadQA test set RO results for three settings: gold QA pairs, gold questions with synthetic answers, and synthetic QA pairs. As documents increase to 256, the gap between synthetic and gold questions narrows with synthetic answers but persists with gold answers. This suggests more data can reduce the gap, but human-annotated answers are still essential for optimal performance.}}
  {\includegraphics[width=1\linewidth]{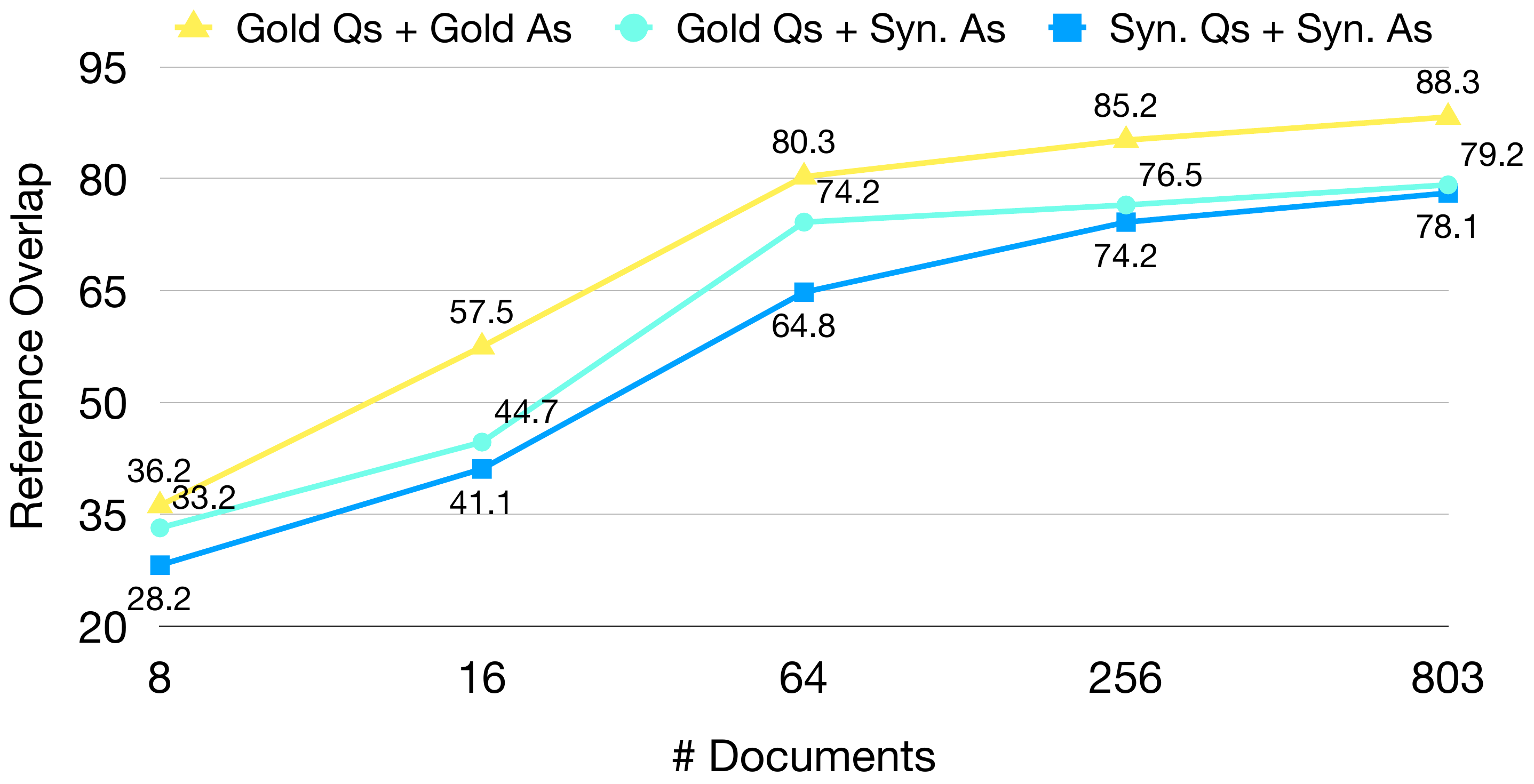}}
\end{figure}

\paragraph{Scaling-up Synthetic Data} 
We explore the impact of scaling up synthetic data by increasing the number of documents from 8 to 803, the full training set size of RadQA. For each document, we generate 5, 10, and 20 QA pairs using ``Summarization + No Overlap" with \texttt{GPT-4o}. Figure \ref{fig:scale-up_ro} shows the test set RO scores of fine-tuned models, with each point representing the average score across three random seeds. Fine-tuned models achieve competitive performance with 64 documents and 5 QA pairs per document. When the number of documents is small (e.g., 8 documents), over-generating QA pairs (20) improves performance. However, as the number of documents increases to 64, generating more QA pairs per document (10 or 20) does not yield additional gains. This suggests that exposing the model to more documents is more beneficial than generating additional questions per document. The best performance of a synthetic data-trained model achieves an RO score of 79.2 with 803 training documents and 5 QA pairs per document. The same trend is observed in the F\textsubscript{1} scores (see Figure \ref{fig:scale-up_f1} in the appendix).

\paragraph{Synthetic v.s. Gold}
We compare synthetic data to gold annotated human data. In addition to directly fine-tuning on gold and synthetic QA pairs, we introduce a new setting where LLMs like \texttt{GPT-4o} generate answers to gold questions, and then fine-tune the model on these pairs. This isolates the impact of answer quality and better assesses the utility of synthetic questions. Similar to the scaling experiments, we fine-tune models on varying document numbers (8 to 803), generating three questions per document using ``Summarization + No Overlap" to match the gold data. Figure \ref{fig:gold_question} shows the RO results. We observe that, with fewer documents (e.g., $\leq 64$), there is a clear gap between synthetic and gold questions regardless of the answer source. As the number of documents increases to 256 or 803, this gap narrows with synthetic answers but persists with gold answers. This suggests that as more data becomes available, the difference between synthetic and gold questions diminishes for model fine-tuning. Nonetheless, the quality of synthetic answers could be a bottleneck for optimal performance. Further limitations of using synthetic Clinical QA data are discussed in Appendix \S\ref{sec:limitations}.

\section{Related Work}

\paragraph{Clinical QA}
Clinical Question Answering systems \citep{PATRICK2012292} aim to deliver precise and reliable answers to medical inquiries derived from electronic health records (EHRs). A significant challenge in developing these systems is the shortage of labeled data. Prior research focuses on curating high-quality datasets \citep{raghavan2018annotating, oliveira2021experiments, fan-2019-annotating, soni-etal-2022-radqa}. However, this approach is not easily scalable due to the substantial costs and time required for manual annotation, especially within the clinical domain where data privacy and security are critical. To address this challenge, \citet{pampari-etal-2018-emrqa} adopt a template-based strategy to construct a large-scale dataset, emrQA. Despite this, the diversity and quality of the generated questions are significantly restricted by the templates. \citet{mimicqa} utilize emrQA to train supervised question generation models, addressing the diversity issue through a question prefix prediction module to encourage different types of questions. Although this method is more adaptable than template-based approaches, it remains constrained by the training data and can only produce answerable questions, which limits its practicality in clinical settings. Our work seeks to overcome these limitations by generating high-quality synthetic data, including unanswerable questions, directly from LLMs using human instructions. This approach enhances scalability and practicality for real-world applications.

\paragraph{QA Corpora Generation}

Numerous prior studies have explored the generation of question-answer pairs \citep{du-etal-2017-learning, alberti-etal-2019-synthetic, puri-etal-2020-training, 10.1162/tacl_a_00415, ushio-etal-2022-generative, ushio-etal-2023-empirical, yoon-bak-2023-diversity}. These methods typically follow a two-stage supervised process: first predicting a candidate answer from an input context, and then generating questions based on these answers. However, their performance is fundamentally constrained by the training data, which limits their applicability to low-resource domains. Efforts have also been made to generate QA corpora in low-resource settings \citep{ram-etal-2021-shot, chada-natarajan-2021-fewshotqa}. For example, \citet{chen2024minprompt} combine template-based question generation with graph-based reasoning to create high-quality synthetic data, using graphs connected through named entities extracted by off-the-shelf tools. Recently, research has increasingly focused on leveraging LLMs for synthetic QA corpora generation. \citet{samuel2023llms} utilize one-shot in-context learning to produce QA pairs. Similar to our zero-shot approach, \citet{viswanathan-etal-2023-prompt2model} generate synthetic data solely from human instructions and improve data diversity through temperature annealing. Despite the potential of these approaches, the generated data are often too simplistic and can be easily solved through superficial matching. Our work addresses these limitations by introducing methods that generate more challenging questions.

\section{Conclusion}


We explore generating Clinical QA data using instruction-tuned LLMs. To address the issue of generating overly simplistic questions with naive prompting, we propose two strategies: 1) instructing LLMs to generate questions that do not overlap with the input context, and 2) summarizing the input context to guide question generation. We demonstrate the effectiveness of our methods on two Clinical QA datasets, consistently outperforming all baselines. Further analyses confirm that our methods generate more challenging questions. Moreover, through carefully designed comparisons, we find that as more data becomes available, the gap between synthetic and gold questions diminishes for model fine-tuning, though human-annotated answers remain essential for optimal performance. 

\acks{This research is supported in part by funding from Optum. The authors would like to thank Optum for their generous support.}

\bibliography{jmlr-sample,custom}

\appendix

\clearpage

\section{Prompts Used}
\label{sec:full_prompt}

\subsection{RadQA Prompts}

\textbf{Direct Instruction} (same for Temp. Anneal.)

\begin{lstlisting}
<radiology_report>
{{input_context}}
</radiology_report>

Considering the radiology report provided above, generate {{question_num}} questions from a medical professional's viewpoint that they would seek to address through a radiological examination, formatted in an indexed list like "1. ... ".
\end{lstlisting}

\noindent\textbf{Question Prefix}

\begin{lstlisting}
<radiology_report>
{{input_context}}
</radiology_report>

Considering the radiology report provided above, generate {{question_num}} questions from a medical professional's viewpoint that they would seek to address through a radiological examination. Please make sure each question starts with a different prefix, such as "is," "does," "has," "which," "what," "how," and "where", formatted in an indexed list like "1. ... ".
\end{lstlisting}

\noindent\textbf{No Overlap}

\begin{lstlisting}
<radiology_report>
{{input_context}}
</radiology_report>

Considering the radiology report provided above, generate {{question_num}} questions from a medical professional's viewpoint that they would seek to address through a radiological examination, formatted in an indexed list like "1. ... ". Make sure that the generated questions do not contain any words from the radiology report.
\end{lstlisting}

\noindent\textbf{Summarization}

\begin{lstlisting}
<radiology_report>
{{input_context}}
</radiology_report>

Output JSON Template:
{
  "symptoms": ["xx"],
  "medical_conditions": ["xx"],
  "areas_examined": ["xx"],
  "patient_medical_history": ["xx"],
  "diagnostic_techniques": ["xx"],
}

Please generate a summary for the radiology report above to cover 5 following aspects: "symptoms", "medical_conditions", "areas_examined", "patient_medical_history", and "diagnostic_techniques", following the JSON template. If there is no information found for an aspect, then just output an empty list [] as the value in the JSON output.
\end{lstlisting}

\noindent\textbf{No Overlap} (after summarization)

\begin{lstlisting}
<patient_data>
{{input_summary}}
</patient_data>

Considering the patient data provided, generate {{question_num}} questions from a medical professional's viewpoint that they would seek to address through a radiological examination. Ensure the questions are diverse, covering various relevant aspects of the patient data. The generated questions should be formatted in an indexed list like "1. ... ". Make sure that the generated questions do not contain any words from the patient data.
\end{lstlisting}

\noindent\textbf{Answer Distillation}

\begin{lstlisting}
<radiology_report>
{{input_context}}
</radiology_report>

Please address the question below by referencing the specific details provided in the preceding report. Employ an extractive question-answering approach: provide only a quotation from the report as the answer, wrapped by quotation marks, and ensure these quotes are as concise as possible to accurately fulfill the query. Make every effort to find the answer in the report, considering all possible details. If, after thorough consideration, the question genuinely cannot be answered with the information provided, respond with "Unanswerable". Always aim to find a relevant and accurate answer. The output should be formatted as "Q: ... <newline>A: ... <newline><newline>Q: ...".

{{input_questions}}
\end{lstlisting}

\subsection{MIMIC-QA Prompts}

\textbf{Direct Instruction} (same for Temp. Anneal.)

\begin{lstlisting}
<clinical_record>
{{input_context}}
</clinical_record>

Considering the clinical record provided above, generate {{question_num}} questions from a medical professional's viewpoint, formatted in an indexed list like "1. ... <newline>2. ...".
\end{lstlisting}

\noindent\textbf{Question Prefix}

\begin{lstlisting}
<clinical_record>
{{input_context}}
</clinical_record>

Considering the clinical record provided above, generate {{question_num}} questions from a medical professional's viewpoint. Please make sure each question starts with a different prefix, such as "is," "does," "has," "which," "what," "how," and "where", formatted in an indexed list like "1. ... ".
\end{lstlisting}

\noindent\textbf{No Overlap}

\begin{lstlisting}
<clinical_record>
{{input_context}}
</clinical_record>

Considering the clinical record provided above, generate {{question_num}} questions from a medical professional's viewpoint, formatted in an indexed list like "1. ... <newline>2. ...". Make sure that the generated questions do not contain any words from the clinical record.
\end{lstlisting}

\noindent\textbf{Summarization}

\begin{lstlisting}
<clinical_record>
{{input_context}}
</clinical_record>

Output JSON Template:
{
    "patient_history": ["value1", "value2", ..., "value5"],
    "diagnosis": ["value1", "value2", ..., "value5"],
    "symptoms": ["value1", "value2", ..., "value5"],
    "medical_conditions": ["value1", "value2", ..., "value5"],
    "exam_results": ["value1", "value2", ..., "value5"],
}

Please generate a structured summary for the clinical record above to cover 5 following aspects: "patient_history", "diagnosis", "symptoms", "medical_conditions", and "exam_results", following the JSON template. Identify five values for each aspect at most. If there is no information found for an aspect, then just output an empty list [] as the value in the JSON output.
\end{lstlisting}

\noindent\textbf{Question Prefix} (after summarization)

\begin{lstlisting}
<patient_data>
{{input_summary}}
</patient_data>

Considering the patient data provided above, generate {{question_num}} questions from a medical professional's viewpoint. Ensure the questions are diverse, covering various relevant aspects of the patient data. Please make sure questions start with different prefixes, such as "is," "does," "has," "which," "what," "how," and "where", formatted in an indexed list like "1. ... ".
\end{lstlisting}

\noindent\textbf{Answer Distillation}

\begin{lstlisting}
<clinical_record>
{{input_context}}
</clinical_record>

Please address the questions below by referencing the specific details provided in the preceding clinical record. Employ an extractive question-answering approach: provide only a quotation from the record as the answer, wrapped by quotation marks. The answer should always be taken from the clinical record and can range from a few words to one or two sentences. For questions beginning with phrases like "does the patient have," "is the patient," etc., ensure the answer is a direct quote from the record rather than a simple yes or no. If, after thorough consideration, the question genuinely cannot be answered with the information provided, respond with "Unanswerable". The output should be formatted as "Q: ... <newline>A: ... <newline><newline>Q: ...".

{{input_questions}}
\end{lstlisting}

\section{Summarization}
\label{sec:att_sum}

\subsection{Summary Example}
\noindent\textbf{Input Document}

\begin{lstlisting}
FINAL REPORT HISTORY: Stroke, G-tube placement, abdominal pain.
Reference exam [**2109-1-18**].
FINDINGS: The G-tube placement is unchanged compared to the prior study with the tip of the pigtail catheter overlying the right upper quadrant. Subcutaneous air is still present. This was a supine film, and therefore, there is limited assessment for free air. Motion somewhat limits the evaluation, but the bowel gas pattern appears grossly normal with stool seen in the colon. There is volume loss in the right lower lobe.
\end{lstlisting}

\noindent\textbf{Generated Summary} (w/ \texttt{GPT-4o})
\begin{lstlisting}
{
  "symptoms": ["abdominal pain"],
  "medical_conditions": ["stroke", "volume loss in the right lower lobe"],
  "areas_examined": ["right upper quadrant", "colon", "right lower lobe"],
  "patient_medical_history": ["stroke", "G-tube placement"],
  "diagnostic_techniques": ["supine film"]
}
\end{lstlisting}

\subsection{Ablations on Summarization Schema}
To demonstrate the importance of the pre-defined summarization schema, which contains high-level attributes of interest to doctors when formulating questions, we conduct an ablation study on RadQA with both \texttt{Llama3-8b} and \texttt{GPT-4o}. We experiment with two new settings for ``Summarization + No Overlap'': 1) Incomplete Schema and 2) No Schema. For ``Incomplete Schema,'' we include only three attributes: ``symptoms,'' ``medical\_conditions,'' and ``patient\_medical\_history,'' leaving out the two radiology-specific attributes, ``areas\_examined'' and ``diagnostic\_techniques.'' For ``No Schema,'' we instruct the model to generate a simple one-paragraph summary and then generate questions based on that summary. We compare these settings with the ``Full Schema,'' with results presented in Table \ref{tab:ablation_schema}. We observe that the ``Full Schema'' consistently outperforms ``No Schema,'' demonstrating the effectiveness of using a high-level schema to guide question generation. The ``Incomplete Schema'' shows comparable performance with \texttt{GPT-4o}, but lower performance with \texttt{Llama3-8b}, indicating that stronger LLMs are more robust to schema design.

\begin{table}[t!]
\centering
\floatconts
    {tab:ablation_schema}
    {\caption{Schema ablation.}}
    {\scalebox{0.7}{
    \begin{tabular}{lcccccccccccc}
    \toprule
     
    & \multicolumn{2}{c}{\texttt{Llama3-8B}} & \multicolumn{2}{c}{\texttt{GPT-4o}} \\
    
    \cmidrule(l){2-3}  \cmidrule(l){4-5}  
    
    \textbf{Method} & F\textsubscript{1} & RO & F\textsubscript{1} & RO \\
    
    \midrule
    
     Full Schema (5 attr.) & \textbf{60.8}\textsubscript{2.3} & \textbf{74.2}\textsubscript{2.3} & \textbf{65.0}\textsubscript{1.2} & \textbf{76.8}\textsubscript{2.9} \\

     Incomplete Schema (3 attr.) & 55.0\textsubscript{0.8} & 66.3\textsubscript{1.3} & 64.0\textsubscript{0.4} & 76.5\textsubscript{1.1} \\

     No Schema & 58.3\textsubscript{1.0} & 71.4\textsubscript{2.1} & 63.9\textsubscript{0.8} & 75.7\textsubscript{0.8} \\

    \bottomrule
    \end{tabular}
    }}
\end{table}

\section{Limitations}
\label{sec:limitations}

Deploying Clinical QA systems trained on synthetic data in real-world clinical settings presents both practical advantages and potential limitations. On the practical side, synthetic data can accelerate the development process by bypassing privacy concerns and data availability issues, enabling rapid prototyping and testing of models. However, synthetic data may not fully capture the complexity and variability of real clinical scenarios. As shown in our analyses in \S\ref{sec:results}, there is a clear difference between synthetic and gold questions in terms of style and question type distribution. Additionally, the quality of the answers hindered the performance of the models trained on synthetic data. Furthermore, ensuring that the synthetic data represents the full spectrum of patient demographics remains a challenge, as biased or incomplete synthetic datasets could lead to suboptimal performance and potential harm in clinical decision-making. Thus, while synthetic data can be a valuable resource for training Clinical QA systems, it should be used more carefully to ensure robustness and generalizability.

Regarding our proposed methods, while designed to be domain-agnostic, they require human-written summarization schemas and data generation prompts. This reliance limits their direct applicability to other clinical sub-domains and general fields such as science \citep{le-etal-2022-shot, bai-etal-2022-synkb} and finance \citep{potluru2024syntheticdataapplicationsfinance}. Future research could focus on leveraging automatic prompt optimization techniques \citep{wang2023promptagent} to minimize human involvement while maintaining the quality of synthetic data.

\begin{figure}[!t]
    \centering
    \includegraphics[width=0.47\textwidth]{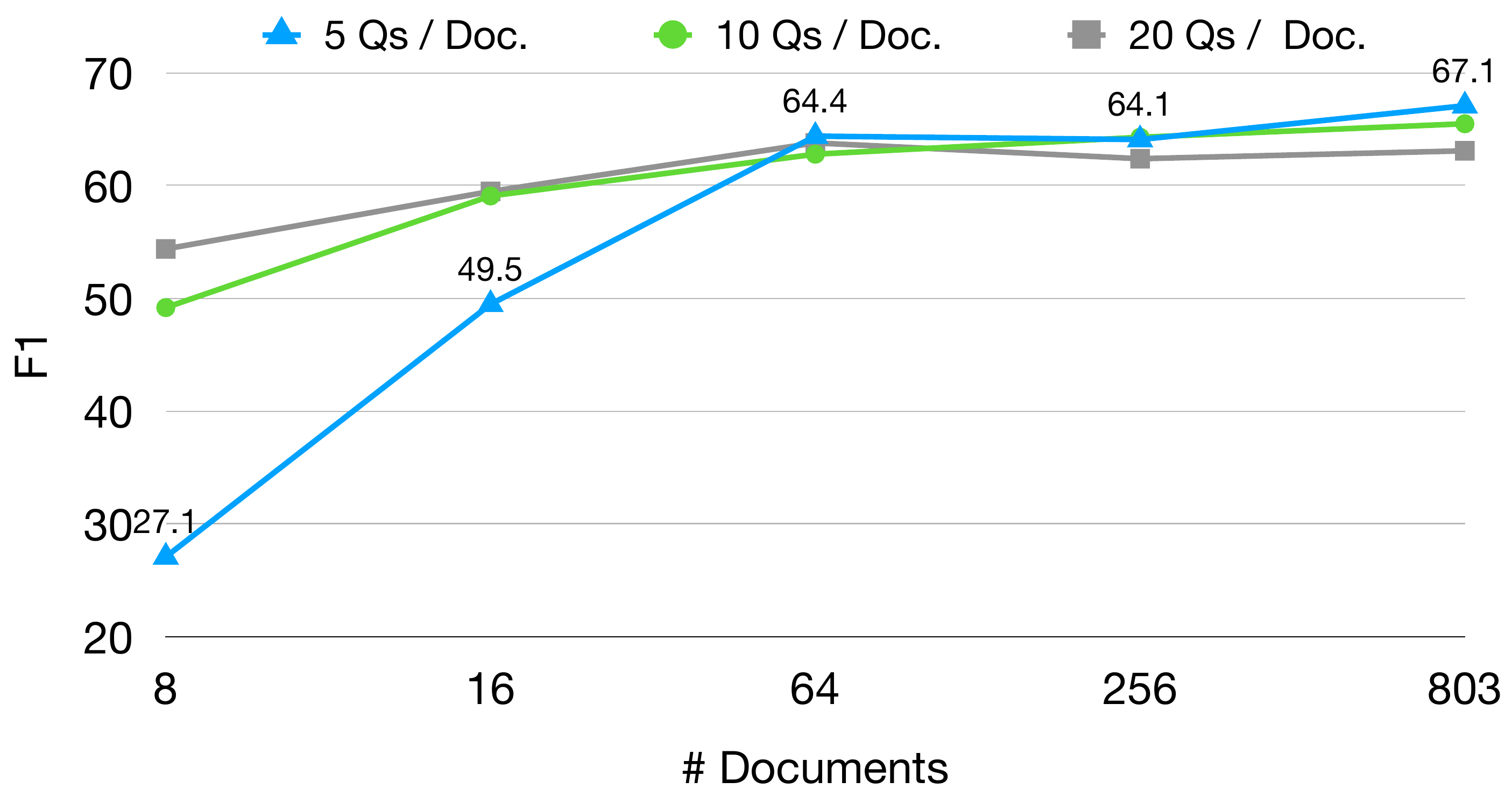}
    \caption{RadQA test set F\textsubscript{1} results from scaling experiments, where the number of documents used is increased from 8 to 803. For each document, we experiment with generating 5, 10, and 20 QA pairs using `` Sum. + No Overlap" with \texttt{GPT-4o}. Similar to the findings from Figure \ref{fig:scale-up_ro}, fine-tuned models achieve competitive performance with 64 documents and 5 QA pairs per document. Increasing the number of QA pairs per document is only helpful in low-resource settings (i.e., 8 or 16 documents).}
    \label{fig:scale-up_f1}
\end{figure}

\begin{table*}[!ht]
\centering
\floatconts
    {tab:main_results}
    {\caption{Test set results on RadQA and MIMIC-QA, where models are fine-tuned on synthetic QA pairs generated from 64 sampled documents in each dataset (640 pairs for RadQA and 845 pairs for MIMIC-QA). Along with Exact Match (EM) and F\textsubscript{1}, we include Reference Overlap (RO), which considers a predicted answer correct if its character-level index in the input document overlaps with the gold answer's index. Results are averaged across three random seeds, with standard deviations indicated as subscripts. Combining ``Summarization" with ``No Overlap" achieves the best performance on RadQA, while ``Summarization" combined with ``Question Prefix" yields the best results on MIMIC-QA.}}
    {\resizebox{1.0\textwidth}{!}{
    \begin{tabular}{lcccccccccccc}
    \toprule
    
     & \multicolumn{6}{c}{\textbf{RadQA}} &  \multicolumn{6}{c}{\textbf{MIMIC-QA}}  \\
    
    \cmidrule(l){2-7}  \cmidrule(l){8-13}
     
    & \multicolumn{3}{c}{\texttt{Llama3-8B}} & \multicolumn{3}{c}{\texttt{GPT-4o}} & \multicolumn{3}{c}{\texttt{Llama3-8B}} & \multicolumn{3}{c}{\texttt{GPT-4o}} \\
    
    \cmidrule(l){2-4}  \cmidrule(l){5-7} \cmidrule(l){8-10} \cmidrule(l){11-13} 
    
    \textbf{Method} & EM & F\textsubscript{1} & RO & EM & F\textsubscript{1} & RO & EM & F\textsubscript{1} & RO & EM & F\textsubscript{1} & RO \\
    
    \midrule
    
    Zero-shot LLM & 37.9 & 59.7 & 70.4 & 46.6 & 64.8 & 74.9 & 23.2 & 53.7 & 48.9 & 24.9 & 58.5 & 66.7 \\
        
    \midrule
    
    Direct Instruct. & 33.5\textsubscript{2.2} & 52.0\textsubscript{1.1} & 61.7\textsubscript{3.7} & 42.0\textsubscript{0.4} & 60.6\textsubscript{1.8} & 70.8\textsubscript{2.7} & 31.8\textsubscript{3.8} & 56.4\textsubscript{4.2} & 58.0\textsubscript{3.5} & 24.3\textsubscript{2.1} & 51.0\textsubscript{1.3} & 54.8\textsubscript{2.9} \\
    
    Temp. Anneal.  & 32.2\textsubscript{0.6} & 51.3\textsubscript{3.5} & 60.9\textsubscript{5.5} & 38.8\textsubscript{2.0} & 55.9\textsubscript{3.7} & 65.1\textsubscript{4.7} & \textbf{33.5}\textsubscript{1.1} & 58.4\textsubscript{2.5} & 59.5\textsubscript{1.4} & 22.5\textsubscript{2.1} & 47.7\textsubscript{5.3} & 51.5\textsubscript{6.4} \\
    
    Ques. Prefix (QP) & 33.3\textsubscript{1.9} & 54.9\textsubscript{1.6} & 65.8\textsubscript{1.6} & 40.6\textsubscript{1.5} & 57.4\textsubscript{1.8} & 66.2\textsubscript{1.3} & 33.2\textsubscript{4.2} & 57.0\textsubscript{4.3} & 57.0\textsubscript{5.4} & 30.4\textsubscript{2.1} & 55.4\textsubscript{0.4} & 57.2\textsubscript{1.3} \\
    
    No Overlap (NO) & 34.3\textsubscript{0.2} & 57.8\textsubscript{1.1} & 70.6\textsubscript{1.5} & \textbf{43.4}\textsubscript{0.9} & 63.7\textsubscript{1.9} & 75.4\textsubscript{2.6} & 23.3\textsubscript{3.7} & 48.8\textsubscript{3.8} & 51.8\textsubscript{2.2} & 25.4\textsubscript{1.6} & 51.6\textsubscript{1.5} & 54.2\textsubscript{3.1}  \\

    Sum. + Direct Instru. & 34.9\textsubscript{0.3} & 55.1\textsubscript{1.3} & 65.5\textsubscript{2.7} & 40.0\textsubscript{1.0} & 59.6\textsubscript{1.3} & 71.1\textsubscript{2.1} & 32.0\textsubscript{0.4} & 56.8\textsubscript{0.6} & 57.1\textsubscript{0.7} & 30.2\textsubscript{2.1} & 54.5\textsubscript{2.9} & 54.6\textsubscript{4.2}\\
    
    Sum. + NO/QP & \textbf{37.6}\textsubscript{2.5} & \textbf{60.8}\textsubscript{2.3} & \textbf{74.2}\textsubscript{2.3} & 43.3\textsubscript{0.9} & \textbf{65.0}\textsubscript{1.2} & \textbf{76.8}\textsubscript{2.9} & 30.8\textsubscript{3.1} & \textbf{58.7}\textsubscript{2.0} & \textbf{61.3}\textsubscript{1.9} & \textbf{31.0}\textsubscript{1.9} & \textbf{57.1}\textsubscript{1.9} & \textbf{58.5}\textsubscript{2.2} \\

    \bottomrule
    \end{tabular}
    }}
\end{table*}

\begin{table*}[th!]
\small
\centering
\floatconts
    {tab:qa_dataset}
    {\caption{Comparison of ten existing Clinical QA datasets. Out of the ten datasets, only four are publicly available (text highlighted in red), and only two, RadQA \citep{soni-etal-2022-radqa} and MIMIC-QA \citep{mimicqa}, involve human-annotated questions and answers (background colored in yellow), which are used in our experiments.
    }}
    
    \scalebox{0.8}{
    
    \begin{tabular}{lcccccccc}
    
    \toprule 

    \multirow{2}{*}{\textbf{Dataset}} & \multirow{2}{*}{\textbf{Availability}} & \multicolumn{3}{c}{\textbf{Annotation}} & \multicolumn{2}{c}{\textbf{Size}} & \textbf{Doc. Source} \\

    \cmidrule(l){3-5} \cmidrule(l){6-7}
    
    & & \textbf{Question} & \textbf{Answer} & \textbf{Unans. Ques.} & \textbf{\# Doc.} & \textbf{\# Ques.} \\

    \midrule
    \textcolor{red}{\textbf{emrQA \citep{pampari-etal-2018-emrqa}}} & \textcolor{red}{\textbf{\ding{51}}} & \textcolor{red}{\textbf{Automatic}} & \textcolor{red}{\textbf{Automatic}} & \textcolor{red}{\textbf{\ding{55}}} & \textcolor{red}{\textbf{303}} & \textcolor{red}{\textbf{73,111}} & \textcolor{red}{\textbf{MIMIC-III}} \\

     \midrule

     \citet{raghavan2018annotating}  & \ding{55} & Manual & Manual & \ding{55} & 71 & 1,747 & Cleveland Clinic \\

    \midrule

    Why-QA \citep{fan-2019-annotating} & \ding{55} & Manual & Manual & \ding{55} & 138 & 245 & i2b2 (n2c2) \\

    \midrule

    \rowcolor{yellow}
    \textcolor{red}{\textbf{MIMIC-QA \citep{mimicqa}}} & \textcolor{red}{\ding{51}} & \textcolor{red}{\textbf{Hybrid}} & \textcolor{red}{\textbf{Hybrid}} & \textcolor{red}{\ding{55}} & \textcolor{red}{\textbf{36}} & \textcolor{red}{\textbf{1,287}} & \textcolor{red}{\textbf{MIMIC-III}} \\

    \midrule

    \citet{oliveira2021experiments}  & \ding{55} & Manual & Manual & \ding{55} & 9 & 18 & SemClinBr
corpus \\

    \midrule

    \textcolor{red}{\textbf{DiSCQ \citep{lehman-etal-2022-learning}}} & \textcolor{red}{\textbf{\ding{51}}} & \textcolor{red}{\textbf{Manual}} & \textcolor{red}{\textbf{-}} & \textcolor{red}{\textbf{\ding{55}}} & \textcolor{red}{\textbf{114}} & \textcolor{red}{\textbf{2,029}} & \textcolor{red}{\textbf{MIMIC-III}} \\

    \midrule

    \rowcolor{yellow}
    \textcolor{red}{\textbf{RadQA \citep{soni-etal-2022-radqa}}} & \textcolor{red}{\textbf{\ding{51}}} & \textcolor{red}{\textbf{Manual}} & \textcolor{red}{\textbf{Manual}} & \textcolor{red}{\ding{51}} & \textcolor{red}{\textbf{1,009}} & \textcolor{red}{\textbf{3,074}} & \textcolor{red}{\textbf{MIMIC-III}} \\

    \midrule

    \citet{mahbub2023}   & \ding{55} & Hybrid & Automatic & \ding{55} & 2,336 & 28,855 & VA CDW \\

    \midrule

    RxWhyQA \citep{rxwhyqa} & \ding{55} & Automatic & Automatic & \ding{51} & 505 & 96,939 & MIMIC-III \\

    \midrule

    \citet{dada2024information}  & \ding{55} & Manual & Manual & \ding{51} & 1,223 & 29,273 & Brain CT Reports \\

     \bottomrule
    \end{tabular}
    }

\end{table*}

\begin{table*}[th!]
\small
\centering
\floatconts
    {tab:examples}
    {\caption{Gold and synthetic QA pairs from each method (w/ \texttt{GPT-4o}) for a sample radiology report. Baseline methods tend to generate questions that closely relate to the context details with lexical overlap (highlighted in bold), while our proposed ``Summarization + No Overlap" generates more challenging questions (introducing relevant but unseen concepts, highlighted in red).
    }}
    {\scalebox{0.95}{
    
    \begin{tabular}{lL{14cm}}
    
    \toprule 
    
    \textbf{Report} & FINAL REPORT HISTORY: Stroke, \textbf{G-tube placement}, abdominal pain. \newline
    Reference exam [**2109-1-18**]. \newline
    FINDINGS: The \textbf{G-tube placement} is unchanged compared to the prior study with the tip of the pigtail catheter overlying the right upper quadrant. \textbf{Subcutaneous air} is still present. This was a supine film, and therefore, there is limited assessment for \textbf{free air}. Motion somewhat limits the evaluation, but the bowel gas pattern appears grossly normal with stool seen in the colon. There is volume loss in the right lower lobe. \\
    
    \midrule
    \midrule
    
    \textbf{\textbf{Method}} & \textbf{\textbf{Gold/Synthetic QA pairs}} \\
    
    \midrule

    \multirow{6}{*}{Gold QA pairs} & \textbf{Q:} Is the \textbf{G tube} patent ? \\
    & \textbf{A:} Unanswerable \\
    & \textbf{Q:} Do we see any \textcolor{red}{pus filled cavities} below the insertion of \textbf{G tube}? \\
    & \textbf{A:} Unanswerable \\
    & \textbf{Q:} Is there a cause for the \textcolor{red}{tenderness} around \textbf{G-Tube} site? \\
    & \textbf{A:} Unanswerable \\

    \midrule 
     
    \multirow{7}{*}{Direct Instruct.} & \textbf{Q:} What is the current position of the \textbf{G-tube}, and has it changed since the last examination? \\
    & \textbf{A:} The G-tube placement is unchanged compared to the prior study \\
    & \textbf{Q:} Is there any evidence of \textbf{free air} in the abdominal cavity, indicating a possible perforation? \\
    & \textbf{A:} therefore, there is limited assessment for free air \\
    & \textbf{Q:} Is the \textbf{subcutaneous air} present in the current film, and has it increased or decreased since the last study? \\
    & \textbf{A:} Subcutaneous air is still present \\
    
     \midrule
    
    \multirow{6}{*}{Temp. Aneal.} & \textbf{Q:} Is the \textbf{G-tube} correctly positioned within the \textcolor{red}{gastrointestinal tract}?\\
    & \textbf{A:} The G-tube placement is unchanged compared to the prior study \\
    & \textbf{Q:} Is there any evidence of \textbf{free air} in the abdomen that might suggest a perforation?\\
    & \textbf{A:} therefore, there is limited assessment for free air \\
    & \textbf{Q:} What is the extent and distribution of the \textbf{subcutaneous air} present? \\
    & \textbf{A:} Subcutaneous air is still present \\
     \midrule
    
    \multirow{7}{*}{Question Prefix} & \textbf{Q:} Is the \textbf{G-tube placement} correct and functioning as intended? \\
    & \textbf{A:} The G-tube placement is unchanged compared to the prior study \\
    & \textbf{Q:} Does the presence of \textbf{subcutaneous air} indicate any complications or issues with the \textbf{G-tube}? \\
    & \textbf{A:} Unanswerable \\
    & \textbf{Q:} Has there been any change in the position of the \textbf{G-tube} compared to the previous study? \\
    & \textbf{A:} The G-tube placement is unchanged compared to the prior study \\
    
     \midrule
    
    \multirow{6}{*}{No Overlap} & \textbf{Q:} Is there any evidence of \textcolor{red}{gastrointestinal perforation}? \\
    & \textbf{A:} limited assessment for free air \\
    & \textbf{Q:} Are there any signs of \textcolor{red}{infection or inflammation} in the \textbf{abdominal} cavity? \\
    & \textbf{A:} Unanswerable \\
    & \textbf{Q:} Is the feeding \textbf{tube} correctly positioned within the \textcolor{red}{stomach or small intestine}?\\
    & \textbf{A:} The G-tube placement is unchanged compared to the prior study \\
    
     \midrule
    
    \multirow{6}{*}{Sum. + No Over.} & \textbf{Q:} Is there any evidence of \textcolor{red}{gastrointestinal perforation or obstruction}?\\
    & \textbf{A:} the bowel gas pattern appears grossly normal \\
    & \textbf{Q:} Are there any signs of \textcolor{red}{inflammation or infection} in the \textbf{abdominal} cavity? \\
    & \textbf{A:} Unanswerable \\
    & \textbf{Q:} Can we identify any abnormalities in the \textcolor{red}{liver or gallbladder region}?\\
    & \textbf{A:} Unanswerable \\
     \bottomrule
    \end{tabular}
    }}

\end{table*}

\end{document}